\documentclass{article} 
\usepackage{iclr2016_conference,times}
\usepackage{hyperref}
\usepackage{url}
\usepackage{graphicx}
\usepackage{amsmath}
\usepackage{amsfonts}
\usepackage{subcaption}
\usepackage{todonotes}

\title{Deep Reinforcement Learning in Parameterized Action Space}
\author{Matthew Hausknecht\\
Department of Computer Science\\
University of Texas at Austin\\
\texttt{mhauskn@cs.utexas.edu}\\
\And
Peter Stone\\
Department of Computer Science\\
University of Texas at Austin\\
\texttt{pstone@cs.utexas.edu}\\
}

\DeclareMathOperator*{\argmax}{arg\,max}

\iclrfinalcopy
\begin{document}
\maketitle
\begin{abstract}
Recent work has shown that deep neural networks are capable of
approximating both value functions and policies in reinforcement
learning domains featuring continuous state and action
spaces. However, to the best of our knowledge no previous work has
succeeded at using deep neural networks in structured (parameterized)
continuous action spaces. To fill this gap, this paper focuses on
learning within the domain of simulated RoboCup soccer, which features
a small set of discrete action types, each of which is parameterized
with continuous variables. The best learned agents can score goals
more reliably than the 2012 RoboCup champion agent. As such, this
paper represents a successful extension of deep reinforcement learning
to the class of parameterized action space MDPs.
\end{abstract}

\section{Introduction}
\label{sec:introduction}

This paper extends the Deep Deterministic Policy Gradients (DDPG)
algorithm \citep{lillicrap15} into a parameterized action space. We
document a modification to the published version of the DDPG
algorithm: namely bounding action space gradients. We found this
modification necessary for stable learning in this domain and will
likely be valuable for future practitioners attempting to learn in
continuous, bounded action spaces.

We demonstrate reliable learning, from scratch, of RoboCup soccer
policies capable of goal scoring. These policies operate on a
low-level continuous state space and a parameterized-continuous action
space. Using a single reward function, the agents learn to locate and
approach the ball, dribble to the goal, and score on an empty
goal. The best learned agent proves more reliable at scoring goals,
though slower, than the hand-coded 2012 RoboCup champion.

RoboCup 2D Half-Field-Offense (HFO) is a research platform for
exploring single agent learning, multi-agent learning, and adhoc
teamwork. HFO features a low-level continuous state space and
parameterized-continuous action space. Specifically, the parameterized
action space requires the agent to first select the type of action it
wishes to perform from a discrete list of high level actions and then
specify the continuous parameters to accompany that action. This
parameterization introduces structure not found in a purely continuous
action space.

The rest of this paper is organized as follows: the HFO domain is
presented in Section \ref{sec:domain}. Section \ref{sec:background}
presents background on deep continuous reinforcement learning
including detailed actor and critic updates. Section
\ref{sec:bounded_action_space} presents a method of bounding action
space gradients. Section \ref{sec:results} covers experiments and
results. Finally, related work is presented in Section
\ref{sec:related} followed by conclusions.

\section{Half Field Offense Domain}
\label{sec:domain}
RoboCup is an international robot soccer competition that promotes
research in AI and robotics. Within RoboCup, the \textit{2D simulation
  league} works with an abstraction of soccer wherein the players, the
ball, and the field are all 2-dimensional objects. However, for the
researcher looking to quickly prototype and evaluate different
algorithms, the full soccer task presents a cumbersome prospect: full
games are lengthy, have high variance in their outcome, and demand
specialized handling of rules such as free kicks and offsides.

The Half Field Offense domain abstracts away the difficulties of full
RoboCup and exposes the experimenter only to core decision-making
logic, and to focus on the most challenging part of a RoboCup 2D game:
scoring and defending goals. In HFO, each agent receives its own state
sensations and must independently select its own actions. HFO is
naturally characterized as an episodic multi-agent POMDP because of
the sequential partial observations and actions on the part of the
agents and the well-defined episodes which culminate in either a goal
being scored or the ball leaving the play area. To begin each episode,
the agent and ball are positioned randomly on the offensive half of
the field. The episode ends when a goal is scored, the ball leaves the
field, or 500 timesteps pass. The following
subsections introduce the low-level state and action space used by
agents in this domain.

\subsection{State Space}
\begin{figure*}[htp]
\begin{subfigure}{.5\textwidth}
  \centering
  \includegraphics[width=0.45\textwidth]{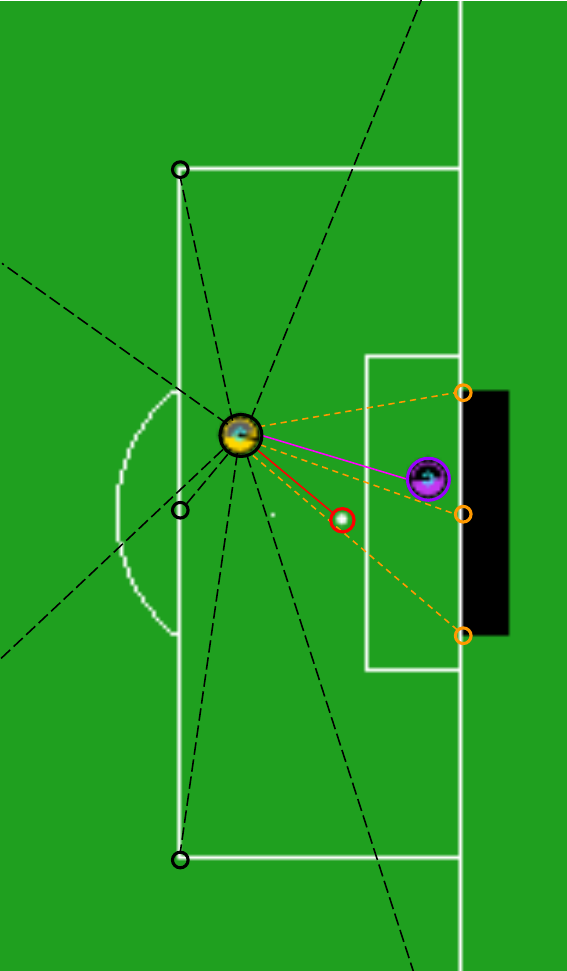}
  \caption{State Space}
\end{subfigure}
\begin{subfigure}{.5\textwidth}
  \centering
  \includegraphics[width=0.6\textwidth]{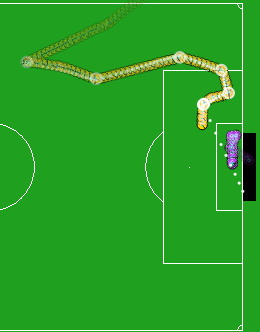}
  \caption{Helios Champion}
\end{subfigure}
\caption{\textbf{Left:} HFO State Representation uses a
  low-level, egocentric viewpoint providing features such as distances
  and angles to objects of interest like the ball, goal posts, corners
  of the field, and opponents. \textbf{Right:} Helios handcoded policy
  scores on a goalie. This 2012 champion agent forms a natural (albeit
  difficult) baseline of comparison.}
\label{fig:state_rep}
\end{figure*}

The agent uses a low-level, egocentric viewpoint encoded using 58
continuously-valued features. These features are derived through
Helios-Agent2D's \citep{akiyama2010agent2d} world model and provide
angles and distances to various on-field objects of importance such as
the ball, the goal, and the other players. Figure \ref{fig:state_rep}
depicts the perceptions of the agent. The most relevant features
include: Agent's position, velocity, and orientation, and stamina;
Indicator if the agent is able to kick; Angles and distances to the
following objects: Ball, Goal, Field-Corners, Penalty-Box-Corners,
Teammates, and Opponents. A full list of state features may be found
at \url{https://github.com/mhauskn/HFO/blob/master/doc/manual.pdf}.

\subsection{Action Space}
\label{subsec:action_space}
Half Field Offense features a low-level, parameterized action
space. There are four mutually-exclusive discrete actions: Dash, Turn,
Tackle, and Kick. At each timestep the agent must select one of these
four to execute. Each action has 1-2 continuously-valued parameters
which must also be specified. An agent must select both the discrete
action it wishes to execute as well as the continuously valued
parameters required by that action. The full set of parameterized
actions is:

\texttt{Dash}(power, direction): Moves in the indicated
direction with a scalar power in $[0,100]$. Movement is faster forward
than sideways or backwards. \texttt{Turn}(direction):
Turns to indicated direction. \texttt{Tackle}(direction):
Contests the ball by moving in the indicated direction. This action is
only useful when playing against an
opponent. \texttt{Kick}(power, direction): Kicks the ball
in the indicated direction with a scalar power in $[0,100]$. All
directions are parameterized in the range of $[-180,180]$ degrees.

\subsection{Reward Signal}
True rewards in the HFO domain come from winning full games. However,
such a reward signal is far too sparse for learning agents to gain
traction. Instead we introduce a hand-crafted reward signal with four
components: \textbf{Move To Ball Reward} provides a scalar reward
proportional to the change in distance between the agent and the ball
$d(a,b)$. An additional reward $\mathbb{I}^{kick}$ of 1 is given the
first time each episode the agent is close enough to kick the
ball. \textbf{Kick To Goal Reward} is proportional to the change in
distance between the ball and the center of the goal $d(b,g)$. An
additional reward is given for scoring a goal $\mathbb{I}^{goal}$. A
weighted sum of these components results in a single reward that first
guides the agent close enough to kick the ball, then rewards for
kicking towards goal, and finally for scoring. It was necessary to
provide a higher gain for the kick-to-goal component of the reward
because immediately following each kick, the move-to-ball component
produces negative rewards as the ball moves away from the agent. The
overall reward is as follows:

\begin{equation}
r_t = d_{t-1}(a,b)-d_t(a,b) + \mathbb{I}^{kick}_t + 3\big(d_{t-1}(b,g)-d_t(b,g)\big) + 5 \mathbb{I}^{goal}_t
\end{equation}

It is disappointing that reward engineering is necessary. However, the
exploration task proves far too difficult to ever gain traction on a
reward that consists only of scoring goals, because acting randomly is
exceedingly unlikely to yield even a single goal in any reasonable
amount of time. An interesting direction for future work is to find
better ways of exploring large state spaces. One recent approach in
this direction, \cite{stadie15} assigned exploration bonuses based on
a model of system dynamics.

\section{Background: Deep Reinforcement Learning}
\label{sec:background}
Deep neural networks are adept general purpose function approximators
that have been most widely used in supervised learning
tasks. Recently, however they have been applied to reinforcement
learning problems, giving rise to the field of deep reinforcement
learning. This field seeks to combine the advances in deep neural
networks with reinforcement learning algorithms to create agents
capable of acting intelligently in complex environments. This section
presents background in deep reinforcement learning in continuous
action spaces. The notation closely follows that of
\cite{lillicrap15}.

Deep, model-free RL in discrete action spaces can be performed using
the Deep Q-Learning method introduced by \cite{mnih15} which employs a
single deep network to estimate the value function of each discrete
action and, when acting, selects the maximally valued output for a
given state input. Several variants of DQN have been explored.
\cite{narasimhan15} used decaying traces, \cite{hausknecht15}
investigated LSTM recurrency, and \cite{hasselt15} explored double
Q-Learning. These networks work well in continuous state spaces but do
not function in continuous action spaces because the output nodes of
the network, while continuous, are trained to output Q-Value estimates
rather than continuous actions.

\begin{figure}[htp]
\centering
\begin{subfigure}{.35\textwidth}
  \centering
  \includegraphics[width=\linewidth]{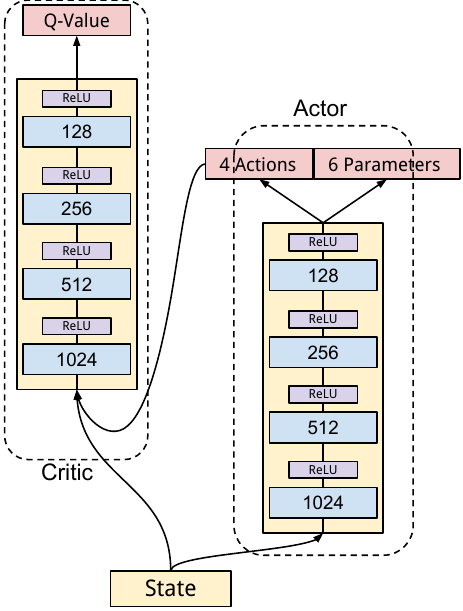}
\end{subfigure}
\hspace{4em}
\begin{subfigure}{.35\textwidth}
  \centering
  \includegraphics[width=\linewidth]{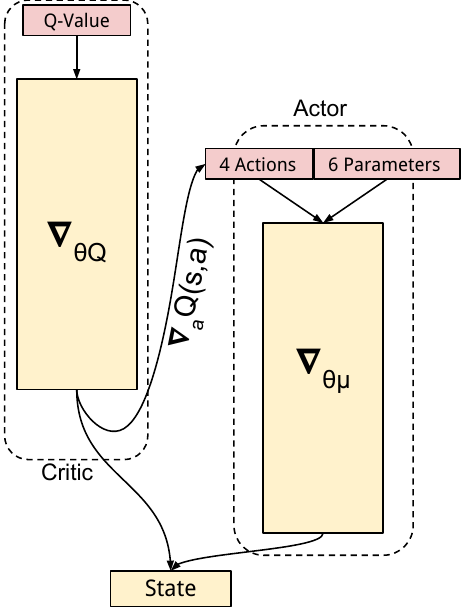}
\end{subfigure}
  \caption{\textbf{Actor-Critic architecture (left)}: actor and critic
    networks may be interlinked, allowing activations to flow forwards
    from the actor to the critic and gradients to flow backwards from
    the critic to the actor. The gradients coming from the critic
    indicate directions of improvement in the continuous action space
    and are used to train the actor network without explicit
    targets. \textbf{Actor Update (right)}: Backwards pass generates
    critic gradients $\nabla_{a}Q(s,a|\theta^Q)$ w.r.t. the
    action. These gradients are back-propagated through the actor
    resulting in gradients w.r.t. parameters $\nabla_{\theta^\mu}$
    which are used to update the actor. Critic gradients
    w.r.t. parameters $\nabla_{\theta^Q}$ are ignored during the actor
    update.}
\label{fig:arch}
\end{figure}

An Actor/Critic architecture \citep{sutton98} provides one solution to
this problem by decoupling the value learning and the action
selection. Represented using two deep neural networks, the actor
network outputs continuous actions while the critic estimates the
value function. The actor network $\mu$, parameterized by
$\theta^\mu$, takes as input a state $s$ and outputs a continuous
action $a$. The critic network $Q$, parameterized by $\theta^Q$, takes
as input a state $s$ and action $a$ and outputs a scalar Q-Value
$Q(s,a)$. Figure \ref{fig:arch} shows Critic and Actor networks.

Updates to the critic network are largely unchanged from the
standard temporal difference update used originally in Q-Learning
\citep{watkins92} and later by DQN:

\begin{equation}
Q(s,a) = Q(s,a) + \alpha \big( r + \gamma \max_{a'} Q(s',a') - Q(s,a) \big)
\end{equation}

Adapting this equation to the neural network setting described above
results in minimizing a loss function defined as follows:

\begin{equation}
L_Q(s,a|\theta^Q) = \Big( Q(s,a|\theta^Q) - \big(r + \gamma \max_{a'} Q(s',a'|\theta^Q)\big) \Big)^2
\end{equation}

However, in continuous action spaces, this equation is no longer
tractable as it involves maximizing over next-state actions
$a'$. Instead we ask the actor network to provide a next-state action
$a' = \mu(s'|\theta^\mu)$. This yields a critic loss with the
following form:

\begin{equation}
L_Q(s,a|\theta^Q) = \Big( Q(s,a|\theta^Q) - \big(r + \gamma Q(s',\mu(s'|\theta^\mu)'|\theta^Q)\big) \Big)^2
\label{eqn:critic_update}
\end{equation}

The value function of the critic can be learned by gradient descent on
this loss function with respect to $\theta^Q$. However, the accuracy
of this value function is highly influenced by the quality of the
actor's policy, since the actor determines the next-state action $a'$
in the update target.

The critic's knowledge of action values is then harnessed to learn a
better policy for the actor. Given a sample state, the goal of the
actor is to minimize the difference between its current output
$a$ and the optimal action in that state $a^*$.

\begin{equation}
L_\mu(s|\theta^\mu) = \big(a - a^* \big)^2 = \big( \mu(s|\theta^Q) - a^* \big)^2
\label{eqn:actor_update}
\end{equation}

The critic may be used to provide estimates of the quality of
different actions but naively estimating $a^*$ would involve
maximizing the critic's output over all possible actions: $a^* \approx
\argmax_a Q(s,a|\theta^Q)$. Instead of seeking a global maximum, the
critic network can provide gradients which indicate directions of
change, in action space, that lead to higher estimated Q-Values:
$\nabla_{a}Q(s,a|\theta^Q)$. To obtain these gradients requires a
single backward pass over the critic network, much faster than solving
an optimization problem in continuous action space. Note that these
gradients are not the common gradients with respect to
parameters. Instead these are gradients with respect to inputs, first
used in this way by NFQCA \citep{hafner11}. To update the actor
network, these gradients are placed at the actor's output layer (in
lieu of targets) and then back-propagated through the network. For a
given state, the actor is run forward to produce an action that the
critic evaluates, and the resulting gradients may be used to update
the actor:

\begin{equation}
\nabla_{\theta^\mu}\mu(s) = \nabla_{a}Q(s,a|\theta^Q)\nabla_{\theta^\mu}\mu(s|\theta^\mu)
\end{equation}

Alternatively one may think about these updates as simply interlinking
the actor and critic networks: On the forward pass, the actor's output
is passed forward into the critic and evaluated. Next, the estimated
Q-Value is backpropagated through the critic, producing gradients
$\nabla_{a}Q$ that indicate how the action should change in order to
increase the Q-Value. On the backwards pass, these gradients flow from
the critic through the actor. An update is then performed only over
the actor's parameters. Figure \ref{fig:arch} shows an example of this
update.

\subsection{Stable Updates}
\label{subsec:stable_updates}
Updates to the critic rely on the assumption that the actor's policy
is a good proxy for the optimal policy. Updates to the actor rest on
the assumption that the critic's gradients, or suggested directions
for policy improvement, are valid when tested in the environment. It
should come as no surprise that several techniques are necessary to
make this learning process stable and convergent.

Because the critic's policy $Q(s,a|\theta^Q)$ influences both the
actor and critic updates, errors in the critic's policy can create
destructive feedback resulting in divergence of the actor, critic, or
both. To resolve this problem \cite{mnih15} introduce a
Target-Q-Network $Q'$, a replica of the critic network that changes on
a slower time scale than the critic. This target network is used to
generate next state targets for the critic update (Equation
\ref{eqn:critic_update}). Similarly a Target-Actor-Network $\mu'$
combats quick changes in the actor's policy.

The second stabilizing influence is a replay memory $\mathcal{D}$, a
FIFO queue consisting of the agent's latest experiences (typically one
million). Updating from mini-batches of experience sampled uniformly
from this memory reduces bias compared to updating exclusively from
the most recent experiences.

Employing these two techniques the critic loss in Equation
\ref{eqn:critic_update} and actor update in Equation
\ref{eqn:actor_update} can be stably re-expressed as follows:

\begin{equation}
L_Q(\theta^Q) = \mathbb{E}_{(s_t,a_t,r_t,s_{t+1}) \sim \mathcal{D}}\bigg[ \Big( Q(s_t,a_t) - \big(r_t + \gamma Q'(s_{t+1},\mu'(s_{t+1}))\big) \Big)^2 \bigg]
\label{eqn:stable_critic_update}
\end{equation}

\begin{equation}
\nabla_{\theta^\mu}\mu = \mathbb{E}_{s_t \sim \mathcal{D}}\bigg[ \nabla_{a}Q(s_t,a|\theta^Q)\nabla_{\theta^\mu}\mu(s_t) |_{a=\mu(s_t)}  \bigg]
\end{equation}

Finally, these updates are applied to the respective networks, where
$\alpha$ is a per-parameter step size determined by the gradient
descent algorithm. Additionally, the target-actor and target-critic
networks are updated to smoothly track the actor and critic using a
factor $\tau \ll 1$:

\begin{equation}
\begin{split}
& \theta^{Q} = \theta^{Q} + \alpha \nabla_{\theta^Q} L_Q(\theta^Q) \\
& \theta^{\mu} = \theta^{\mu} + \alpha \nabla_{\theta^\mu} \mu \\
& \theta^{Q'} = \tau \theta^{Q} + (1 - \tau)\theta^{Q'} \\
& \theta^{\mu'} = \tau \theta^{\mu} + (1 - \tau)\theta^{\mu'}
\end{split}
\end{equation}

One final component is an adaptive learning rate method such as
ADADELTA \citep{zeiler12}, RMSPROP \citep{tieleman12}, or ADAM
\citep{ba14}.

\subsection{Network Architecture}
\label{sec:arch}
Shown in Figure \ref{fig:arch}, both the actor and critic employ the
same architecture: The 58 state inputs are processed by four fully
connected layers consisting of 1024-512-256-128 units
respectively. Each fully connected layer is followed by a rectified
linear (ReLU) activation function with negative slope
$10^{-2}$. Weights of the fully connected layers use Gaussian
initialization with a standard deviation of $10^{-2}$. Connected to
the final inner product layer are two linear output layers: one for
the four discrete actions and another for the six parameters
accompanying these actions. In addition to the 58 state features, the
critic also takes as input the four discrete actions and six action
parameters. It outputs a single scalar Q-value. We use the ADAM solver
with both actor and critic learning rate set to $10^{-3}$. Target
networks track the actor and critic using a $\tau = 10^{-4}$. Complete
source code for our agent is available at
\url{https://github.com/mhauskn/dqn-hfo} and for the HFO domain
at \url{https://github.com/mhauskn/HFO/}. Having introduced the
background of deep reinforcement learning in continuous action space,
we now present the parameterized action space.

\section{Parameterized Action Space Architecture}
\label{subsec:parameterized_arch}
Following notation in \citep{masson15}, a Parameterized Action Space
Markov Decision Process (PAMDP) is defined by a set of discrete
actions $A_d = \{a_1,a_2,\dots,a_k\}$. Each discrete action $a \in
A_d$ features $m_a$ continuous parameters $\{p^a_1,\dots,p^a_{m_a}\}
\in \mathbb{R}^{m_a}$. Actions are represented by tuples
$(a,p^a_1,\dots,p^a_{m_a})$. Thus the overall action space $A =
\cup_{a \in A_d}(a, p^a_1, \dots, p^a_{m_a})$.

In Half Field Offense, the complete parameterized action space
(Section \ref{subsec:action_space}) is $A = (\textrm{Dash},
p_1^\textrm{dash}, p_2^\textrm{dash}) \cup (\textrm{Turn},
p_3^\textrm{turn}) \cup (\textrm{Tackle}, p_4^\textrm{tackle}) \cup
(\textrm{Kick}, p_5^\textrm{kick}, p_6^\textrm{kick})$. The actor
network in Figure \ref{fig:arch} factors the action space into one
output layer for discrete actions $(\textrm{Dash}, \textrm{Turn},
\textrm{Tackle}, \textrm{Kick})$ and another for all six continuous
parameters $(p_1^\textrm{dash}, p_2^\textrm{dash}, p_3^\textrm{turn},
p_4^\textrm{tackle}, p_5^\textrm{kick}, p_6^\textrm{kick})$.

\subsection{Action Selection and Exploration}
Using the factored action space, deterministic action selection
proceeds as follows: At each timestep, the actor network outputs
values for each of the four discrete actions as well as six continuous
parameters. The discrete action is chosen to be the maximally valued
output $a = \max(\textrm{Dash}, \textrm{Turn}, \textrm{Tackle},
\textrm{Kick})$ and paired with associated parameters from the
parameter output layer $(a, p^a_1, \dots, p^a_{m_a})$. Thus the actor
network simultaneously chooses which discrete action to execute and
how to parameterize that action.

During training, the critic network receives, as input, the values of
the output nodes of all four discrete actions and all six action
parameters. We do not indicate to the critic which discrete action was
actually applied in the HFO environment or which continuous parameters
are associated with that discrete action. Similarly, when updating the
actor, the critic provides gradients for all four discrete actions and
all six continuous parameters. While it may seem that the critic is
lacking crucial information about the structure of the action space,
our experimental results in Section \ref{sec:results} demonstrate that
the critic learns to provide gradients to the correct parameters of
each discrete action.

Exploration in continuous action space differs from discrete space. We
adapt $\epsilon$-greedy exploration to parameterized action space:
with probability $\epsilon$, a random discrete action $a \in A_d$ is
selected and the associated continuous parameters
$\{p^a_1,\dots,p^a_{m_a}\}$ are sampled using a uniform random
distribution. Experimentally, we anneal $\epsilon$ from 1.0 to 0.1
over the first $10,000$ updates. \cite{lillicrap15} demonstrate that
Ornstein-Uhlenbeck exploration is also successful in continuous action
space.



\section{Bounded Parameter Space Learning}
\label{sec:bounded_action_space}
The Half Field Offense domain bounds the range of each continuous
parameter. Parameters indicating direction (e.g. Turn and Kick
direction) are bounded in $[-180,180]$ and parameters for power
(e.g. Kick and Dash power) are bounded in $[0,100]$. Without enforcing
these bounds, after a few hundred updates, we observed continuous
parameters routinely exceeding the bounds. If updates were permitted
to continue, parameters would quickly trend towards astronomically
large values. This problem stems from the critic providing gradients
that encourage the actor network to continue increasing a parameter
that already exceeds bounds. We explore three approaches for
preserving parameters in their intended ranges:

\textbf{Zeroing Gradients}: Perhaps the simplest approach is to
examine the critic's gradients for each parameter and zero the
gradients that suggest increasing/decreasing the value of a parameter
that is already at the upper/lower limit of its range:

\begin{equation}
  \nabla_{p} =
  \begin{cases}
    \nabla_{p} &\text{if $p_{\min} < p < p_{\max}$} \\
    0 &\text{otherwise}
  \end{cases}
\end{equation}

Where $\nabla_{p}$ indicates the critic's gradient with respect to
parameter $p$, (e.g. $\nabla_{p}Q(s_t,a|\theta^Q)$) and $p_{\min},
p_{\max}, p$ indicate respectively the minimum bound, maximum bound,
and current activation of that parameter.

\textbf{Squashing Gradients}: A squashing function such as the
hyperbolic tangent (tanh) is used to bound the activation of each
parameter. Subsequently, the parameters are re-scaled into their
intended ranges. This approach has the advantage of not requiring
manual gradient tinkering, but presents issues if the squashing
function saturates.

\textbf{Inverting Gradients}: This approach captures the best aspects
of the zeroing and squashing gradients, while minimizing the
drawbacks. Gradients are downscaled as the parameter approaches the
boundaries of its range and are inverted if the parameter exceeds the
value range. This approach actively keeps parameters within bounds
while avoiding problems of saturation. For example, if the critic
continually recommends increasing a parameter, it will converge to the
parameter's upper bound. If the critic then decides to decrease that
parameter, it will decrease immediately. In contrast, a squashing
function would be saturated at the upper bound of the range and
require many updates to decrease. Mathematically, the inverted
gradient approach may be expressed as follows:

\begin{equation}
  \nabla_{p} = \nabla_{p} \cdot
  \begin{cases}
    (p_{\max} - p) / (p_{\max} - p_{\min}) &\text{if $\nabla_{p}$ suggests increasing $p$} \\
    (p - p_{\min}) / (p_{\max} - p_{\min}) &\text{otherwise}
  \end{cases}
\end{equation}

It should be noted that these approaches are not specific to HFO or
parameterized action space. Any domain featuring a bounded-continuous
action space will require a similar approach for enforcing bounds. All
three approaches are empirically evaluated the next section.

\section{Results}
\label{sec:results}
We evaluate the zeroing, squashing, and inverting gradient approaches
in the parameterized HFO domain on the task of approaching the ball
and scoring a goal. For each approach, we independently train two
agents. All agents are trained for 3 million iterations, approximately
20,000 episodes of play. Training each agent took three days on a
NVidia Titan-X GPU.

Of the three approaches, only the inverting gradient shows robust
learning. Indeed both inverting gradient agents learned to reliably
approach the ball and score goals. None of the other four agents using
the squashing or zeroing gradients were able to reliably approach the
ball or score.

Further analysis of the squashing gradient approach reveals that
parameters stayed within their bounds, but squashing functions quickly
became saturated. The resulting agents take the same discrete action
with the same maximum/minimum parameters each timestep. Given the
observed proclivity of the critic's gradients to push parameters
towards ever larger/small values, it is no surprise that squashing
function quickly become saturated and never recover.

Further analysis of the zeroing gradient approach reveals two
problems: 1) parameters still overflow their bounds and 2)
instability: While the gradient zeroing approach negates any direct
attempts to increase a parameter $p$ beyond its bounds, we hypothesize
the first problem stems from gradients applied to other parameters
$p_i \neq p$ which inadvertently allow parameter $p$ to
overflow. Empirically, we observed learned networks attempting to dash
with a power of 120, more than the maximum of 100. It is reasonable
for a critic network to encourage the actor to dash faster.

\begin{figure*}[htp]
\begin{subfigure}{.33\textwidth}
  \centering
  \includegraphics[width=\textwidth]{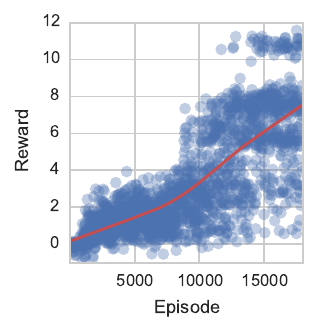}
\end{subfigure}
\begin{subfigure}{.33\textwidth}
  \centering
  \includegraphics[width=\textwidth]{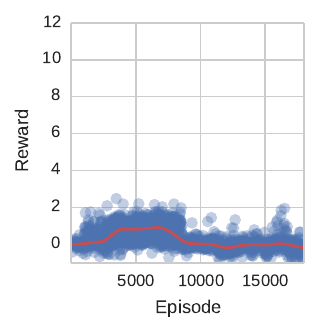}
\end{subfigure}
\begin{subfigure}{.33\textwidth}
  \centering
  \includegraphics[width=\textwidth]{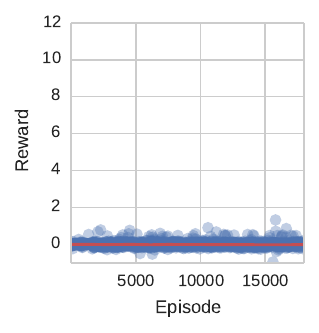}
\end{subfigure}
\begin{subfigure}{.33\textwidth}
  \centering
  \includegraphics[width=\textwidth]{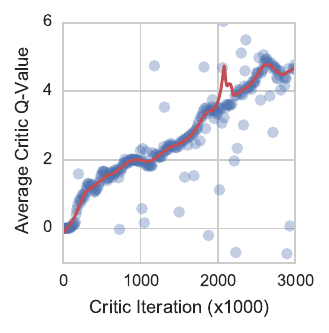}
\end{subfigure}
\begin{subfigure}{.33\textwidth}
  \centering
  \includegraphics[width=\textwidth]{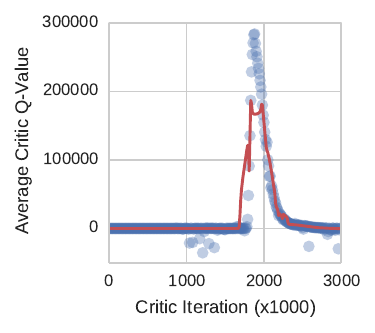}
\end{subfigure}
\begin{subfigure}{.33\textwidth}
  \centering
  \includegraphics[width=\textwidth]{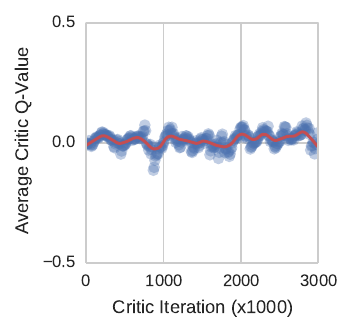}
\end{subfigure}
\begin{subfigure}{.33\textwidth}
  \centering
  \includegraphics[width=\textwidth]{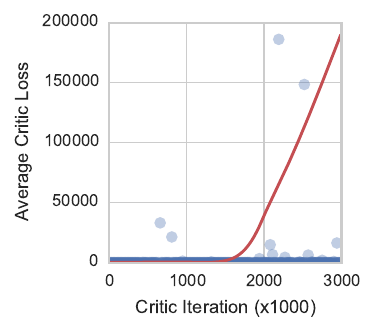}
  \caption{Inverting Gradients}
\end{subfigure}
\begin{subfigure}{.33\textwidth}
  \centering
  \includegraphics[width=\textwidth]{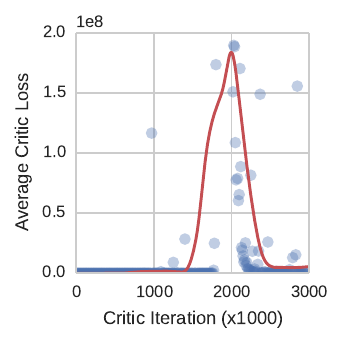}
  \caption{Zeroing Gradients}
\end{subfigure}
\begin{subfigure}{.33\textwidth}
  \centering
  \includegraphics[width=\textwidth]{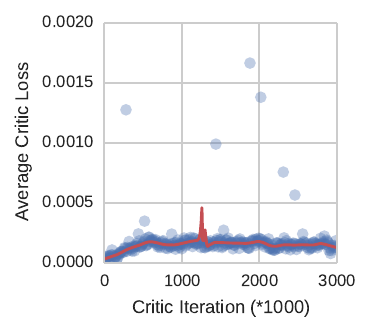}
  \caption{Squashing Gradients}
\end{subfigure}
\caption{\textbf{Analysis of gradient bounding strategies:} The
  left/middle/right columns respectively correspond to the
  inverting/zeroing/squashing gradients approaches to handling bounded
  continuous actions. \textbf{First row} depicts learning curves
  showing overall task performance: Only the inverting gradient
  approach succeeds in learning the soccer task. \textbf{Second row}
  shows average Q-Values produced by the critic throughout the entire
  learning process: Inverting gradient approach shows smoothly
  increasing Q-Values. The zeroing approach shows astronomically high
  Q-Values indicating instability in the critic. The squashing
  approach shows stable Q-Values that accurately reflect the actor's
  performance. \textbf{Third row} shows the average loss experienced
  during a critic update (Equation \ref{eqn:stable_critic_update}): As
  more reward is experienced critic loss is expected to rise as past
  actions are seen as increasingly sub-optimal. Inverting gradients
  shows growing critic loss with outliers accounting for the rapid
  increase nearing the right edge of the graph. Zeroing gradients
  approach shows unstably large loss. Squashing gradients never
  discovers much reward and loss stays near zero.}
\label{fig:analysis}
\end{figure*}

Unstable learning was observed in one of the two zeroing gradient
agents. This instability is well captured in the Q-Values and critic
losses shown in Figure \ref{fig:analysis}. It's not clear why this
agent became unstable, but the remaining stable agent showed clear
results of not learning.

These results highlight the necessity of non-saturating functions that
effectively enforce action bounds. The approach of inverting gradients
was observed to respect parameter boundaries (observed dash power
reaches 98.8 out of 100) without saturating. As a result, the critic
was able to effectively shape the actor's policy. Further evaluation
of the reliability and quality of the inverting-gradient policies is
presented in the next section.

\section{Soccer Evaluation}
\label{sec:eval}
We further evaluate the inverting gradient agents by comparing them to
an expert agent independently created by the Helios RoboCup-2D
team. This agent won the 2012 RoboCup-2D world championship and source
code was subsequently released \citep{akiyama2010agent2d}. Thus, this
hand-coded policy represents an extremely competent player and a high
performance bar.

As an additional baseline we compare to a SARSA learning
agent. State-Action-Reward-State-Action (SARSA) is an algorithm for
model-free on-policy Reinforcement Learning \cite{sutton98}. The SARSA
agent learns in a simplified version of HFO featuring high-level
discrete actions for moving, dribbling, and shooting the ball. As
input it is given continuous features that including the distance and
angle to the goal center. Tile coding \cite{sutton98} is used to
discretize the state space. Experiences collected by playing the game
are then used to bootstrap a value function.

To show that the deep reinforcement learning process is reliable, in
additional to the previous two inverting-gradient agents we
independently train another five inverting-gradient agents, for a
total of seven agents DDPG$_{1-7}$. All seven agents learned to score
goals. Comparing against the Helios' champion agent, each of the
learned agents is evaluated for 100 episodes on how quickly and
reliably it can score.

Six of seven DDPG agents outperform the SARSA baseline, and
remarkably, three of the seven DDPG agents score more reliably than
Helios' champion agent. Occasional failures of the Helios agent result
from noise in the action space, which occasionally causes missed
kicks. In contrast, DDPG agents learn to take extra time to score each
goal, and become more accurate as a result. This extra time is
reasonable considering DDPG is rewarded only for scoring and
experiences no real pressure to score more quickly. We are encouraged
to see that deep reinforcement learning can produce agents competitive
with and even exceeding an expert handcoded agent.

\begin{figure*}[htp]
\begin{subfigure}{.4\textwidth}
  \centering
  \includegraphics[width=\textwidth]{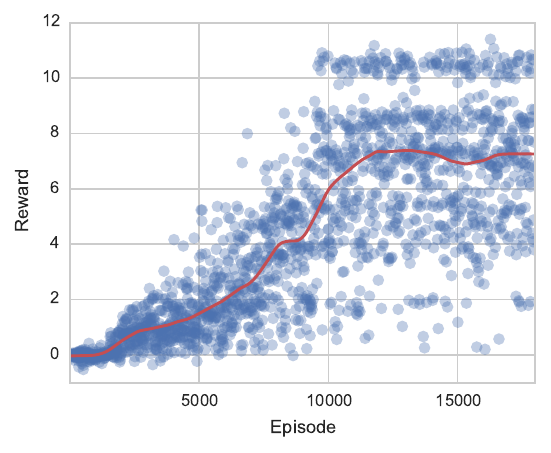}
  \caption{Learning Curve}
\end{subfigure}
\begin{subfigure}{.5\textwidth}
  \centering
  \begin{tabular}{ c | c c }
    & Scoring & Avg. Steps \\
    & Percent & to Goal \\
    \hline
    Helios' Champion & .962 & 72.0 \\
    SARSA & .81 & 70.7 \\
    DDPG$_1$ & 1 & 108.0 \\
    DDPG$_2$ & .99 & 107.1 \\
    DDPG$_3$ & .98 & 104.8 \\
    DDPG$_4$ & .96 & 112.3 \\
    DDPG$_5$ & .94 & 119.1 \\
    DDPG$_6$ & .84 & 113.2 \\
    DDPG$_7$ & .80 & 118.2 \\
  \end{tabular}
  \caption{Evaluation Performance}
\end{subfigure}
\caption{\textbf{Left:} Scatter plot of learning curves of DDPG-agents
  with Lowess curve. Three distinct phases of learning may be seen:
  the agents first get small rewards for approaching the ball (episode
  1500), then learn to kick the ball towards the goal (episodes 2,000
  - 8,000), and start scoring goals around episode
  10,000. \textbf{Right:} DDPG-agents score nearly as reliably as
  expert baseline, but take longer to do so. A video of DDPG$_1$'s
  policy may be viewed at \url{https://youtu.be/Ln0Cl-jE_40}.}
\label{fig:learning}
\end{figure*}

\section{Related Work}
\label{sec:related}
RoboCup 2D soccer has a rich history of learning. In one of the
earliest examples, \cite{andre99} used Genetic Programming to evolve
policies for RoboCup 2D Soccer. By using a sequence of reward
functions, they first encourage the players to approach the ball, kick
the ball, score a goal, and finally to win the game. Similarly, our
work features players whose policies are entirely trained and have no
hand-coded components. Our work differs by using a gradient-based
learning method paired with using reinforcement learning rather than
evolution.

\cite{masson15} present a parameterized-action MDP formulation and
approaches for model-free reinforcement learning in such
environments. The core of this approach uses a parameterized policy
for choosing which discrete action to select and another policy for
selecting continuous parameters for that action. Given a fixed policy
for parameter selection, they use Q-Learning to optimize the policy
discrete action selection. Next, they fix the policy for discrete
action selection and use a policy search method to optimize the
parameter selection. Alternating these two learning phases yields
convergence to either a local or global optimum depending on whether
the policy search procedure can guarantee optimality. In contrast, our
approach to learning in parameterized action space features a
parameterized actor that learns both discrete actions and parameters
and a parameterized critic that learns only the action-value
function. Instead of relying on an external policy search procedure,
we are able to directly query the critic for gradients. Finally, we
parameterize our policies using deep neural networks rather than
linear function approximation. Deep networks offer no theoretical
convergence guarantees, but have a strong record of empirical success.

Experimentally, \cite{masson15} examine a simplified abstraction of
RoboCup 2D soccer which co-locates the agent and ball at the start of
every trial and features a smaller action space consisting only of
parameterized kick actions. However, they do examine the more
difficult task of scoring on a keeper. Since their domain is
hand-crafted and closed-source, it's hard to estimate how difficult
their task is compared to the goal scoring task in our paper.

Competitive RoboCup agents are primarily handcoded but may feature
components that are learned or optimized. \cite{AAAI15-MacAlpine2}
employed the layered-learning framework to incrementally learn a
series of interdependent behaviors. Perhaps the best example of
comprehensively integrating learning is the Brainstormers who, in
competition, use a neural network to make a large portion of decisions
spanning low level skills through high level strategy
\citep{riedmiller09,riedmiller07}. However their work was done prior to
the advent of deep reinforcement learning, and thus required more
constrained, focused training environments for each of their
skills. In contrast, our study learns to approach the ball, kick
towards the goal, and score, all within the context of a single,
monolithic policy.


Deep learning methods have proven useful in various control
domains. As previously mentioned DQN \citep{mnih15} and DDPG
\citep{lillicrap15} provide great starting points for learning in
discrete and continuous action spaces. Additionally, \cite{levine15}
demonstrates the ability of deep learning paired with guided policy
search to learn manipulation policies on a physical robot. The high
requirement for data (in the form of experience) is a hurdle for
applying deep reinforcement learning directly onto robotic
platforms. Our work differs by examining an action space with latent
structure and parameterized-continuous actions.

\section{Future Work}
\label{sec:future_work}
The harder task of scoring on a goalie is left for future
work. Additionally, the RoboCup domain presents many opportunities for
multi-agent collaboration both in an adhoc-teamwork setting (in which
a single learning agent must collaborate with unknown teammates) and
true multi-agent settings (in which multiple learning agents must
collaborate). Challenges in multi-agent learning in the RoboCup domain
have been examined by prior work \citep{shivaram06} and solutions may
translate into the deep reinforcement learning settings as
well. Progress in this direction could eventually result in a team of
deep reinforcement learning soccer players.

Another interesting possibility is utilizing the critic's gradients
with respect to state inputs $\nabla_{s}Q(s,a|\theta^Q)$. These
gradients indicate directions of improvement in state space. An agent
with a forward model may be able to exploit these gradients to
transition into states which the critic finds more favorable. Recent
developments in model-based deep reinforcement learning \citep{oh15}
show that detailed next state models are possible.

\section{Conclusion}
\label{sec:conclusion}
This paper has presented an agent trained exclusively with deep
reinforcement learning which learns from scratch how to approach the
ball, kick the ball to goal, and score. The best learned agent scores
goals more reliably than a handcoded expert policy. Our work does not
address more challenging tasks such as scoring on a goalie or
cooperating with a team, but still represents a step towards fully
learning complex RoboCup agents. More generally we have demonstrated
the capability of deep reinforcement learning in parameterized action
space.

To make this possible, we extended the DDPG algorithm
\citep{lillicrap15}, by presenting an analyzing a novel approach for
bounding the action space gradients suggested by the Critic. This
extension is not specific to the HFO domain and will likely
prove useful for any continuous, bounded action space.

\subsubsection*{Acknowledgments}
The authors wish to thank Yilun Chen. This work has taken place in the
Learning Agents Research Group (LARG) at the Artificial Intelligence
Laboratory, The University of Texas at Austin.  LARG research is
supported in part by grants from the National Science Foundation
(CNS-1330072, CNS-1305287), ONR (21C184-01), AFRL (FA8750-14-1-0070),
AFOSR (FA9550-14-1-0087), and Yujin Robot. Additional support from the
Texas Advanced Computing Center, and Nvidia Corporation.

\bibliography{dqn-hfo}
\bibliographystyle{iclr2016_conference}

\end{document}